\documentclass[conference]{IEEEtran}
\IEEEoverridecommandlockouts
\usepackage[utf8]{inputenc}
\usepackage{cite}
\usepackage{float}
\usepackage[numbers]{natbib}
\usepackage{amsmath,amssymb,amsfonts}

\usepackage{graphicx}
\usepackage{textcomp}
\usepackage{xcolor}
\usepackage{tabularx}
\usepackage{booktabs}
\usepackage{subcaption}
\usepackage{scrextend}
\usepackage{threeparttable}
\usepackage{fancyhdr}
\usepackage{lipsum,float}
\usepackage{booktabs}
\usepackage{multirow}
\usepackage{tabularx}
\usepackage{makecell}
\usepackage{array}
\usepackage{algorithm}
\usepackage{algpseudocode}
\usepackage{booktabs}  
\usepackage{array}
\usepackage{tabularx}
\usepackage{hyperref}
\usepackage{afterpage}
\usepackage{cuted}
\usepackage{graphicx}
\usepackage{stfloats}

\hypersetup{
    colorlinks=true,
    linkcolor=black,      
    citecolor=black,
    urlcolor=black
}

\def\BibTeX{{\rm B\kern-.05em{\sc i\kern-.025em b}\kern-.08em
    T\kern-.1667em\lower.7ex\hbox{E}\kern-.125emX}}

\makeatletter
\renewcommand\subsubsection{%
  \@startsection{subsubsection}{3}{\z@}
    {1.0ex plus 0.5ex minus 0.2ex}
    {0.5ex plus 0.1ex}
    {\normalfont\normalsize\itshape}
}
\makeatother

\usepackage{graphicx}   

\begin{document}

\title{DyFuLM: An Advanced Multimodal Framework for Sentiment Analysis\\
}
\makeatletter
\newcommand{\linebreakand}{\end{@IEEEauthorhalign}\hfill\mbox{}\par
\mbox{}\hfill\begin{@IEEEauthorhalign}}
\makeatother

\author{

\IEEEauthorblockN{
Ruohan Zhou\textsuperscript{1},
Jiachen Yuan\textsuperscript{2},
Churui Yang\textsuperscript{3},
Wenzheng Huang\textsuperscript{2},
Guoyan Zhang\textsuperscript{1},\\
Shiyao Wei\textsuperscript{1},
Jiazhen Hu\textsuperscript{2},
Ning Xin\textsuperscript{2},
Md Maruf Hasan\textsuperscript{*2}
}
\hspace*{-2.5cm}
\IEEEauthorblockA{
\textsuperscript{1}\textit{Department of Applied Mathematics, Xi'an Jiaotong-Liverpool University}, Suzhou, 215123, Jiangsu, China\\
\textsuperscript{2}\textit{School of AI and Advanced Computing, XJTLU Entrepreneur College (Taicang)}, \\Xi'an Jiaotong-Liverpool University, Suzhou, 215123, Jiangsu, China\\
\textsuperscript{3}\textit{Department of Intelligent Science, Xi'an Jiaotong-Liverpool University}, Suzhou, 215123, Jiangsu, China\\
Ruohan.Zhou22@student.xjtlu.edu.cn, Jiachen.Yuan23@student.xjtlu.edu.cn,
Churui.Yang24@student.xjtlu.edu.cn,\\
Wenzheng.Huang23@student.xjtlu.edu.cn, Guoyan.Zhang23@student.xjtlu.edu.cn,
Shiyao.Wei22@student.xjtlu.edu.cn,\\
Jiazhen.Hu24@student.xjtlu.edu.cn, Ning.Xin21@student.xjtlu.edu.cn, MdMaruf.Hasan@xjtlu.edu.cn
}
}

\maketitle

\begin{abstract}
 Understanding sentiment in complex textual expressions remains a fundamental challenge in affective computing. To address this, we propose a Dynamic Fusion Learning Model (DyFuLM), a multimodal framework designed to capture both hierarchical semantic representations and fine-grained emotional nuances. DyFuLM introduces two key modules: a Hierarchical Dynamic Fusion module that adaptively integrates multi-level features, and a Gated Feature Aggregation module that regulates cross-layer information flow to achieve balanced representation learning. Comprehensive experiments on multi-task sentiment datasets demonstrate that DyFuLM achieves 82.64\% coarse-grained and 68.48\% fine-grained accuracy, yielding the lowest regression errors (MAE = 0.0674, MSE = 0.0082) and the highest coefficient of determination ($R^2$ = 0.6903). Furthermore, the ablation study validates the effectiveness of each module in DyFuLM. When all modules are removed, the accuracy drops by 0.91\% for coarse-grained and 0.68\% for fine-grained tasks. Keeping only the gated fusion module causes decreases of 0.75\% and 0.55\%, while removing the dynamic loss mechanism results in drops of 0.78\% and 0.26\% for coarse-grained and fine-grained sentiment classification, respectively. These results demonstrate that each module contributes significantly to feature interaction and task balance. Overall, the experimental findings further validate that DyFuLM enhances sentiment representation and overall performance through effective hierarchical feature fusion.
\end{abstract}


\section{Introduction}
With the rapid growth of the digital economy, hotel reviews on social media and online platforms have become a key factor influencing tourists' decisions \cite{b1}. Traditional methods, such as manual questionnaires, rating-based evaluations, and statistical summaries, provide limited interpretability and fail to capture the implicit emotions and contextual subtleties embedded in travelers' feedback. Compared to traditional methods, the ratings offer insight into perceptions of services and environments. Analyzing the rating of such reviews supports a better understanding of user behavior, enables service optimization, and improves tourism demand forecasting \cite{b2}. Sentiment analysis has been widely adopted in tourism research, helping to allocate resources and personalized recommendations \cite{b3}.

Conventional sentiment analysis methods typically classify text as positive, negative, or neutral based on sentiment lexicons, handcrafted statistical features, or traditional classification algorithms \cite{b4}. However, these approaches often overlook subtle or mixed emotions within the same review, for example, when a user praises the service of a hotel but criticizes its facilities \cite{b5, b6, b7}. The emergence of deep learning has transformed sentiment analysis, with large language models (LLMs) such as BERT exhibiting exceptional capabilities in contextual understanding and semantic representation \cite{b8}. Compared to traditional methods, models such as BERT and RoBERTa achieve higher accuracy and robustness in sentiment classification \cite{b9, b10}. Nevertheless, two significant challenges persist: (1) model performance varies across text types and domains due to differences in vocabulary and linguistic expression, which reduces accuracy \cite{b11}; (2) most existing models still rely on coarse sentiment polarity classification (positive or negative), lacking the ability to quantify emotional intensity \cite{b12}.

To address the above challenges, we propose a Dynamic Fusion Learning Model (DyFuLM). The model integrates disentangled attention to achieve fine-grained semantic representation and improve contextual dependency to enhance robustness. By a multimodal framework, DyFuLM effectively improves emotion recognition in complex and mixed review scenarios. Evaluated on a large-scale dataset containing 515,738 reviews from 1,493 European luxury hotels collected between 2015 and 2017, DyFuLM demonstrates strong performance in sentiment classification tasks. To further enhance domain adaptability, we conduct fine-tuning within the target domain using authentic hotel reviews, which improves both accuracy and robustness of the model. Moreover, DyFuLM employs a multimodal framework that integrates coarse-grained sentiment classification, fine-grained emotion categorization, and emotion intensity regression, enabling the model to capture subtle and multifaceted emotional expressions with greater precision.

The main research contributions of this study are:
\begin{enumerate}
  \item[1.]We propose a multimodal framework that integrates global semantic information with local emotional cues to achieve comprehensive contextual modeling and fine-grained feature representation. The framework jointly performs coarse-grained classification, fine-grained emotion categorization, and emotion intensity regression within a unified architecture. This joint optimization strategy effectively reduces task bias and enhances the model’s generalization ability across multiple emotional dimensions;

  \item[2.]To further refine feature interaction, we introduce a multimodal framework that spans both encoders. This module utilizes gating and adaptive weighting to facilitate cross-layer and cross-model semantic alignment, thereby enhancing representational accuracy;

\end{enumerate}

\section{Related Work}
Sentiment analysis has progressed through three main stages. Early methods relied on traditional machine learning algorithms such as Naïve Bayes, Support Vector Machines \cite{b13}, and Decision Trees \cite{b14}, which extracted handcrafted features using TF–IDF or N-gram representations but failed to capture contextual dependencies \cite{b14}. With the advent of neural networks, CNNs were introduced for local feature extraction and LSTMs for sequential modeling \cite{b15,b16}. However, these architectures required large labeled datasets and suffered from gradient vanishing over long sequences. The emergence of Transformer-based models, particularly BERT and its variants (RoBERTa \cite{b17}, DeBERTa \cite{b18}), marked a major paradigm shift by leveraging self attention and large-scale pretraining to achieve state-of-the-art (SOTA) performance across NLP tasks. Despite these advances, single encoder architectures still face inherent limitations in large scale sentiment analysis. Their unified parameter space restricts representational diversity, leading to domain overfitting. Relying solely on final layer features overlooks valuable linguistic information from intermediate layers, while shared representations in multi-task learning often cause semantic interference across tasks. To overcome these limitations, recent studies have explored multi-encoder frameworks. Poly-Encoder and TwinBERT employ dual-tower architectures for efficient semantic matching but perform fusion only at the final layer, limiting cross layer interaction \cite{b19,b20}. MoE-BERT introduces expert routing for dynamic feature selection, yet its discrete gating mechanism lacks fine-grained semantic alignment \cite{b21}. Similarly, concatenation based and attention-weighted fusion strategies remain static and struggle to adaptively integrate cross model information. To address these challenges, we propose DyFuLM, a multimodal framework learning model with three key innovations: (1) a hierarchical dual-encoder architecture that enhances contextual representation through global–local semantic complementarity, (2) a BiLSTM-guided dynamic fusion mechanism enabling adaptive cross-layer feature extraction, and (3) a gated aggregation module that facilitates fine-grained cross model interaction under hierarchical multi-task guidance. Together, these innovations enable comprehensive and fine-grained sentiment modeling in large scale scenarios.

\section{Methodology}
This section presents the overall methodology of our study, comprising four core components: Data Preprocessing, Workflow Design, Model Architecture, and Architectural Comparisons. Each component outlines the key procedures and design principles that collectively underpin the DyFuLM framework.

\subsection{Data Preprocessing}
\begin{figure}[ht]
  \centering
  \includegraphics[width=1.0\linewidth]{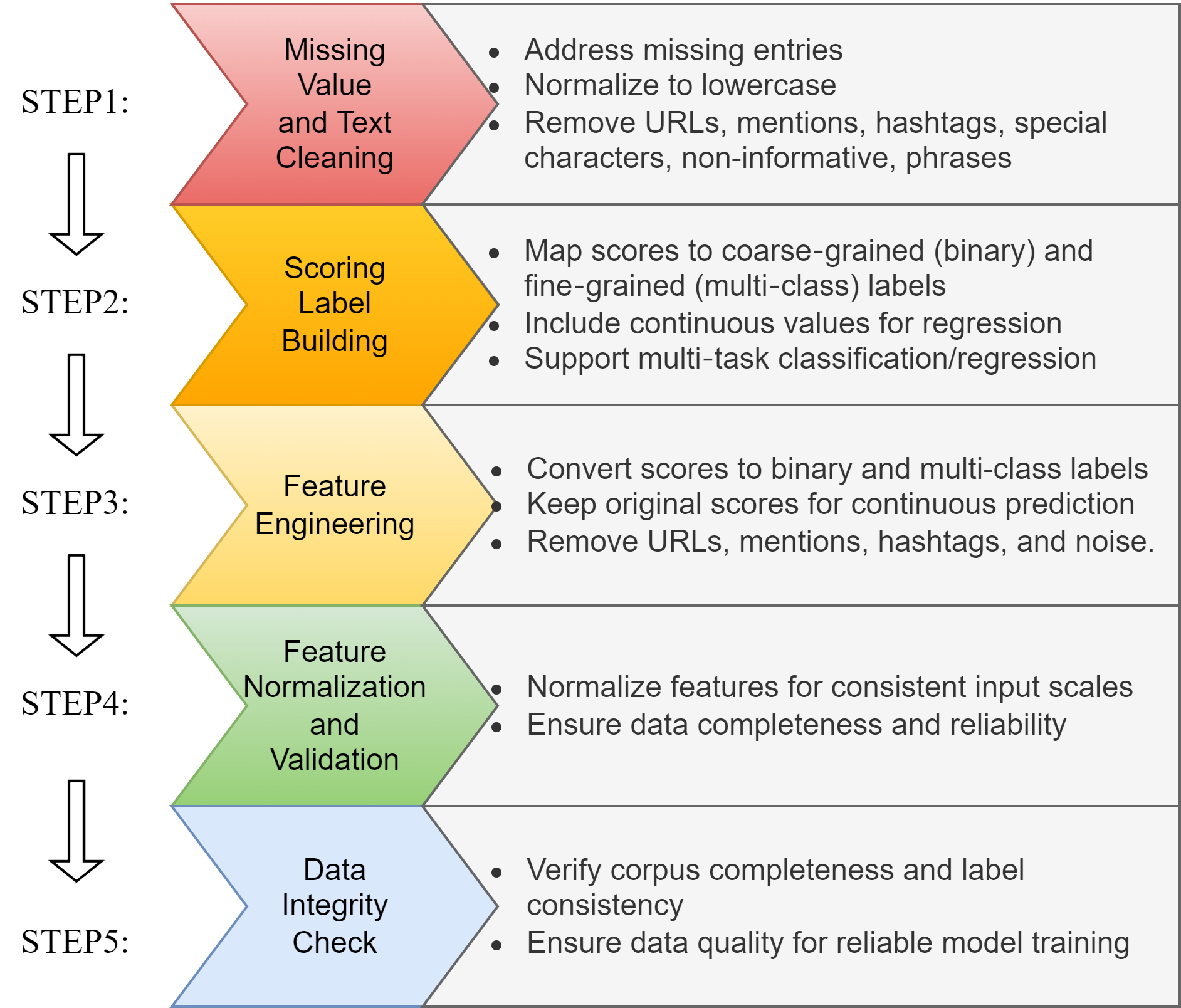}
  \caption{Flowchart of Data Preprocessing}
  \label{fig1}
\end{figure}

To ensure a clean and structured dataset for downstream tasks, the preprocessing procedure involved the Fig.\ref{fig1} and the following steps:

\begin{enumerate}
  \item \textbf{Missing Value and Text Cleaning:} Missing entries were appropriately addressed. Texts were normalized by lowercasing and cleaned by removing URLs, mentions, hashtags, special characters, extra whitespace, and non‑informative phrases.
  \item \textbf{Scoring Label Building:} Reviewer scores were mapped to sentiment labels using both coarse‑grained and fine‑grained schemes to support flexible classification tasks.
  \item \textbf{Feature Engineering:} Key features such as review length and temporal trends were extracted. Sentiment and length distributions were visualized, and geographic patterns were analyzed.
  \item \textbf{Feature Normalization and Validation:} Textual and auxiliary features were transformed and normalized to ensure consistent scales among input variables. This procedure contributed to stable model training and better generalization performance.
  \item \textbf{Data Integrity Check:} A final verification was conducted to confirm corpus completeness, label consistency, and data reliability, ensuring that the dataset was suitable for subsequent model development and evaluation.
\end{enumerate}

\begin{figure*}[!t]
    \centering
    \includegraphics[width=1\textwidth]{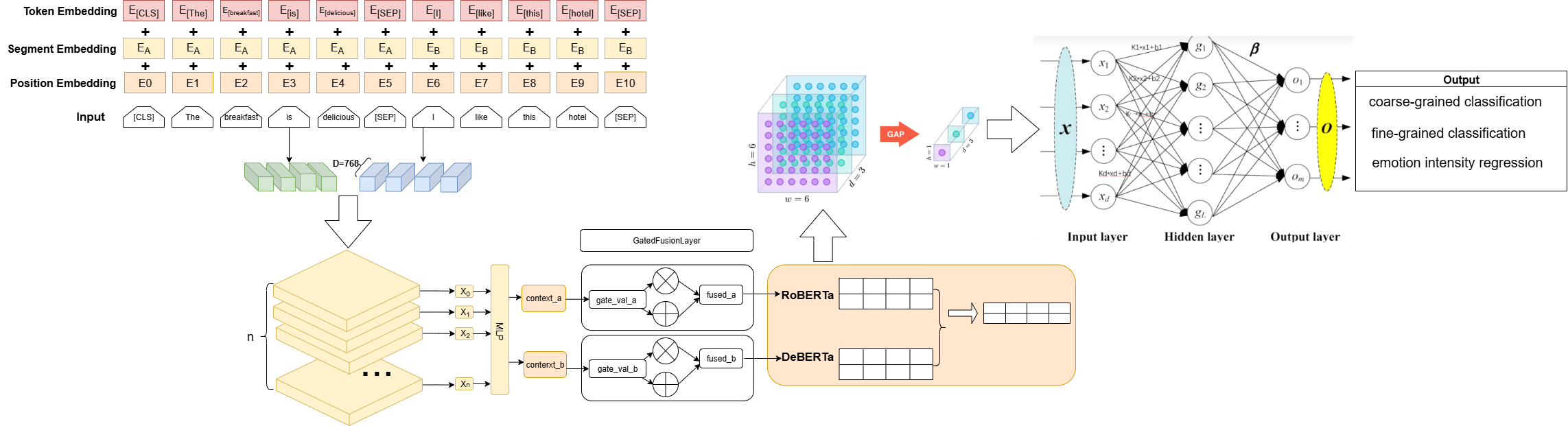}
    \caption{DyFuLM Workflow}
    \label{fig2}
\end{figure*}

\subsection{Workflow}

Fig.\ref{fig2} delineates a multimodal emotion analysis framework. Initially, the preprocessed text is encoded through token, segment, and position embeddings to obtain the basic semantic representations. Subsequently, feature extraction is performed through a stack of layers, and the outputs are concatenated thereafter. A dynamic fusion feature processor, equipped with a gating mechanism, is then employed. This mechanism comprises two branches content\_a and content\_b, each integrating gate\_val and load\_val operations to dynamically regulate feature flow and fusion. Following this, features from  multimodal are processed, followed by Global Average Pooling (GAP). Ultimately, a neural network consisting of input, hidden, and output layers yields results for coarse-grained classification, fine-grained classification, and emotion intensity regression.

\subsection{Model Architecture}
Traditional multi-task sentiment analysis models often rely on a single language encoder for semantic representation. While this structure provides effective global semantic understanding, it struggles to capture detailed emotional cues due to its limited representational depth. To overcome this limitation, we design a multimodal framework collaborative architecture that jointly models contextual semantics and fine-grained emotional features, improving both representation quality and feature integration.

\noindent\textbf{Multimodal Framework Collaborative Hybrid Framework}

DyFuLM introduces a cooperative multimodal framework design that enables dynamic interaction between different encoders. Specifically, RoBERTa contributes strong global contextual understanding and holistic sentiment representation, while DeBERTa provides finer-grained semantic discrimination and captures localized emotional cues. This collaborative integration allows DyFuLM to model both coarse and fine semantic structures more effectively, leading to improved downstream performance.

\noindent\textbf{Hierarchical Dynamic Feature Fusion}

Building upon the multimodal framework backbone, we further introduce a hierarchical dynamic feature fusion mechanism to leverage the diverse information captured at different layers of Transformers. Relying solely on the final layer may overlook informative intermediate representations. To address this, we design a layer-wise attention weighting strategy, allowing the model to dynamically select and aggregate features from multiple layers based on specific task needs. This enables a more flexible and comprehensive integration of multi-level semantics.

To capture inter-layer dependencies, we denote the hidden state of the $l$-th layer as $H^{(l)} \in \mathbb{R}^{T \times d}$, where $T$ is the sequence length and $d$ is the hidden dimension. 
A bidirectional LSTM is first applied to encode the representations from all layers, producing contextualized layer features:
Eq.\ref{eq1}\cite{b15}:
\begin{equation}\label{eq1}
U^{(l)} = \mathrm{BiLSTM}(H^{(1)}, H^{(2)}, \ldots, H^{(L)}).
\end{equation}
To adaptively integrate multi-layer information, a learnable attention weight $\alpha^{(l)}$ is assigned to each layer. 
After softmax normalization, the contribution of each layer is computed as:
Eq.\ref{eq2}\cite{b13}:
\begin{equation}\label{eq2}
\alpha^{(l)} = \frac{\exp(w^{\top} U^{(l)})}{\sum_{k=1}^{L} \exp(w^{\top} U^{(k)})}.
\end{equation}
Here, $L$ denotes the total number of Transformer layers, and $w$ is a learnable weight vector used to compute the attention score for each layer.
The final fused representation is then obtained through a weighted sum:
Eq.\ref{eq3}\
\begin{equation}\label{eq3}
H_{\text{fused}} = \sum_{l=1}^{L} \alpha^{(l)} H^{(l)}.
\end{equation}
This layer-wise attention mechanism allows the model to automatically focus on the most informative layers for each token, enhancing hierarchical feature representation.

\noindent\textbf{Gating-based Cross-model Fusion}

To enhance coordination between models, we introduce a gating mechanism to selectively integrate information between the representations of multimodal framework.

This mechanism acts as a controllable switch that adaptively determines, for each token, whether its final representation should primarily rely on its own contextual features or incorporate complementary cues from the other model. 
Such an adaptive strategy avoids redundant information caused by simple concatenation and promotes more effective feature fusion.

Formally, let the token representation from RoBERTa be denoted as $h_A$, and the DeBERTa representation incorporating contextual information as $c_B$. 
The gating function is defined as:
Eq.\ref{eq4}\
\begin{equation}\label{eq4}
g = \sigma(W [h_A; c_B] + b),
\end{equation}
where $\sigma(\cdot)$ denotes the sigmoid function, and $W$ and $b$ are learnable parameters. 
The operator $[\,\cdot\,; \cdot\,]$ indicates vector concatenation. 
The final fused representation is computed as:
Eq.\ref{eq5}\
\begin{equation}\label{eq5}
h_{\text{fused}} = g \odot h_A + (1 - g) \odot c_B.
\end{equation}
This gating design enables the model to dynamically control the contribution of each source, achieving fine-grained and adaptive information fusion.

\noindent\textbf{Multitasking Output Branches}

To achieve multi-level task modeling, we design three independent yet hierarchically connected prediction heads. 
First, the coarse-grained head $f_{\text{coarse}}$ takes the fused feature representation $h$ as input and produces the coarse-level classification result:
Eq.\ref{eq6}\
\begin{equation}\label{eq6}
\hat{y}_c = f_{\text{coarse}}(h).
\end{equation}
Then, the intensity head $f_{\text{intensity}}$ estimates the emotional strength from the same feature:
Eq.\ref{eq7}\
\begin{equation}\label{eq7}
\hat{y}_i = f_{\text{intensity}}(h).
\end{equation}
Based on these outputs, a guidance function $g = Guidance(\hat{y}_c, \hat{y}_i)$ integrates both predictions to generate a guiding signal that recalibrates the original feature representation:
Eq.\ref{eq8}\
\begin{equation}\label{eq8}
h' = h \odot g,
\end{equation}
where $\odot$ denotes multiplication of corresponding elements.
Finally, the fine-grained head $f_{\text{fine}}$ takes the recalibrated feature $h'$ as input and produces the refined prediction:
Eq.\ref{eq9}\
\begin{equation}\label{eq9}
\hat{y}_f = f_{\text{fine}}(h').
\end{equation}
This hierarchical prediction mechanism from top to bottom allows the coarse classification and intensity modeling to mutually reinforce each other, enhancing both the model’s representational capacity and predictive accuracy.

Here, $h$ denotes the fused feature vector, $\hat{y}_c$ the coarse-level prediction, $\hat{y}_i$ the intensity prediction, $g$ the guidance signal, and $\hat{y}_f$ the final fine-grained prediction.

\subsection{Comparison of Model architectures}

\begin{figure}[H]
  \centering
  \includegraphics[width=1.0\linewidth]{figure2.4.png}
 \caption{Comparison between BERT and DyFuLM architectures (green denotes shared components, red denotes DyFuLM-specific modules)}
 \label{fig3}
\end{figure}

DyFuLM adopts several key enhancements. Fig.\ref{fig3} illustrates the overall architecture, where DyFuLM integrates RoBERTa and DeBERTa as multimodal framework to enable parallel representation learning. To effectively combine their multi-layer features, a hierarchical dynamic fusion mechanism is employed. At the output stage, the model performs three tasks: coarse-grained classification, fine-grained classification, and emotion intensity regression, thereby supporting multi-dimensional sentiment modeling. This design preserves semantic understanding while substantially improving the model’s granularity and generalization in emotion recognition.

\section{Experiments Results and Analysis}
\subsection{Multivariate Time Series Data Display}
This study utilizes a dataset of 515,738 user reviews from Booking.com, encompassing 1,493 hotels across Europe and collected between 2015 and 2017. Each record contains 17 attributes, including review text, rating, timestamp, user nationality, and hotel location. To minimize the influence of outliers and ensure consistency, the original review texts and ratings are directly used as standardized inputs for subsequent analysis.

\begin{table}[ht]
  \centering
  \caption{Statistical characteristics of hotel ratings}
  \label{tab1}
  \small
  \begin{tabularx}{\linewidth}{l *{4}{>{\centering\arraybackslash}X}}
    \hline
    Indicator                   & Maximum Value & 25\% Quartile & Mean   & Median \\ 
    \hline
    2015 Reviewer\_Score        & 10.0000       & 7.5000        & 8.3198 & 8.8000 \\ 
    2016 Reviewer\_Score        & 10.0000       & 7.5000        & 8.4247 & 8.8000 \\ 
    2017 Reviewer\_Score        & 10.0000       & 7.5000        & 8.3905 & 8.8000 \\ 
    Overall Reviewer\_Score     & 10.0000       & 7.5000        & 8.3951 & 8.8000 \\ 
    \hline
  \end{tabularx}
\end{table}
Table \ref{tab1} shows that user ratings for European hotels remained consistently high from 2015 to 2017, with a stable median of 8.8 and a 25th percentile of 7.5 across all three years, indicating a generally positive user perception. The average rating slightly fluctuated, with values of 8.32 in 2015, 8.42 in 2016, and 8.39 in 2017, yet remained at a high level overall. The maximum rating consistently reached 10, confirming a stable scoring mechanism. As shown in Table \ref{tab1}, the number of reviews peaked in 2016, marking a notable increase from 2015, followed by a slight decline in 2017. This fluctuation may be attributed to changes in platform activity or data collection policies.

\begin{figure}[ht]
  \centering
  \includegraphics[width=1.0\linewidth]{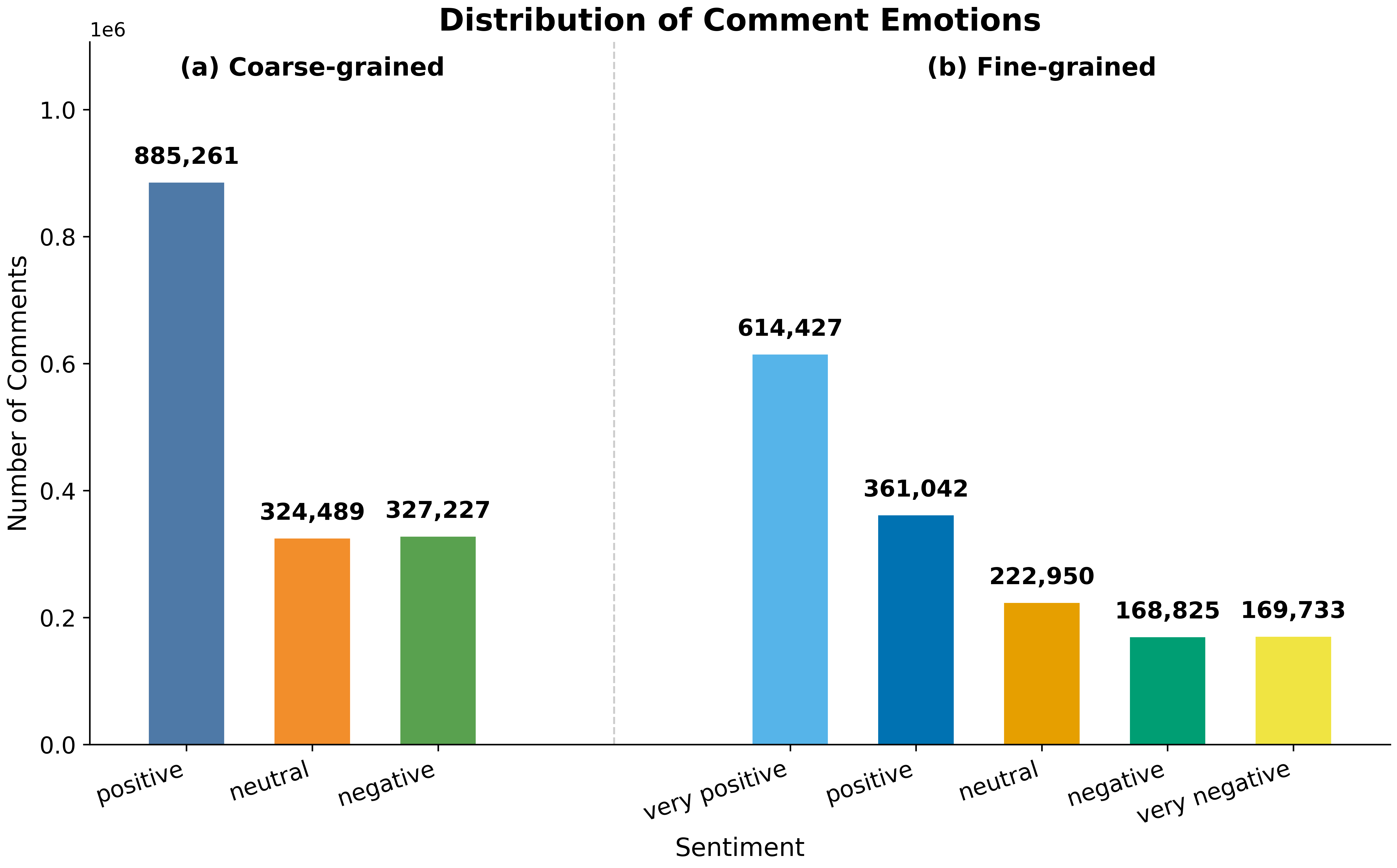}
 \caption{Emotion Category Distribution of Comments}
 \label{fig4}
\end{figure}

Fig.\ref{fig4} presents the fine-grained and coarse-grained sentiment distributions of the reviews. The fine-grained classification divides sentiments into five levels, allowing a more nuanced capture of users’ emotional variations. Among these, positive reviews (very positive and positive) exceed 975,469, forming the majority, while negative reviews total around 338,558. In contrast, the coarse-grained classification simplifies sentiment into three categories, offering a more intuitive overview. It shows 885,261 positive reviews, though at the cost of compressing sentiment detail. The fine-grained approach provides a clearer distinction in sentiment intensity, for example, differentiating between “very positive” and “positive” or between “slightly negative” and “strongly negative,” which is essential for modeling subtle emotional trends in user feedback.

\subsection{Performance Metrics Overview}
Table \ref{tab2} presents a performance comparison between DyFuLM and several representative baselines. We evaluate the models across multiple dimensions. Specifically, Coarse Accuracy and Fine Accuracy measure classification performance at different granularities, while Coarse F1 and Fine F1 assess the balance between precision and recall. Additionally, we report the Mean Absolute Error (MAE) and Mean Squared Error (MSE) for regression tasks, where lower values indicate better prediction accuracy. We also include the $R^2$ (coefficient of determination), which reflects the proportion of variance explained by the model.

\subsection{Performance Comparison and Improvement Analysis}

To benchmark our model, we evaluate it against two widely adopted baseline families.

Single Models. We select five representative models as baselines to ensure a comprehensive comparison across different Transformer developments. BERT serves as the standard baseline in NLP tasks \cite{b8}. RoBERTa, a widely adopted model that achieved state-of-the-art(SOTA) performance upon its release in 2019, reflects improvements in large scale pretraining  \cite{b17}. DeBERTa-v3 represents advanced architectural optimization for contextual understanding \cite{b18}. DistilBERT serves as an efficiency oriented baseline that balances accuracy and speed \cite{b22}. Finally, EmoBERTa, fine-tuned for emotion recognition, provides a domain specific comparison point \cite{b23}.

Hybrid Architectures. To ensure fair and representative evaluation, we include six widely used hybrid baselines that integrate pretrained Transformers with lightweight sequence learners or feature fusion schemes. These baselines are selected because they reflect mainstream strategies for enhancing contextual representations in sentiment analysis. Specifically, Simple + Concat and Gated Fusion represent typical feature fusion mechanisms \cite{b24,b25}, while BERT/RoBERTa + CNN, + BiLSTM, and + GRU correspond to common Transformer encoder integration paradigms \cite{b26,b27,b28,b29}. Together, these baselines provide a comprehensive benchmar for evaluating the effectiveness of our approach. 

\begin{table}[ht]
\centering
\caption {Performance comparison with baselines (Values are presented to four decimal places, with optimal results for each metric in bold.)}
\label{tab2}
\small
\resizebox{1.0\columnwidth}{!}{
\begin{tabular}{lccccccc}
\toprule
\textbf{Model} & \textbf{C. Acc} & \textbf{F. Acc} & \textbf{C. F1} & \textbf{F. F1} & \textbf{MAE} & \textbf{MSE} & \textbf{R²} \\
\midrule
\textbf{DyFuLM} & \textbf{0.8264} & \textbf{0.6848} & \textbf{0.8215} & \textbf{0.6814} & \textbf{0.0674} & \textbf{0.0082} & \textbf{0.6903} \\
\midrule
\textit{Single-Model} & & & & & & & \\
BERT & 0.8128 & 0.6730 & 0.8041 & 0.6656 & 0.0718 & 0.0097 & 0.6366 \\
RoBERT & 0.8133 & 0.6734 & 0.8089 & 0.6692 & 0.0701 & 0.0098 & 0.6360 \\
DeBERT & 0.8163 & 0.6774 & 0.8109 & 0.6704 & 0.0700 & 0.0094 & 0.6485 \\
DistilBERT & 0.8116 & 0.6708 & 0.8031 & 0.6633 & 0.0717 & 0.0098 & 0.6327 \\
EmoBERT & 0.8118 & 0.6670 & 0.8063 & 0.6547 & 0.0764 & 0.0100 & 0.6282 \\
\midrule
\textit{Dual-Architecture} & & & & & & & \\
Simple+Concat & 0.8149 & 0.6732 & 0.8114 & 0.6690 & 0.0714 & 0.0097 & 0.6367 \\
Gated Fusion & 0.8152 & 0.6744 & 0.8097 & 0.6718 & 0.0717 & 0.0098 & 0.6355 \\
RoBERT+CNN & 0.8142 & 0.6731 & 0.8055 & 0.6624 & 0.0733 & 0.0096 & 0.6430 \\
RoBERT+BiLSTM & 0.8130 & 0.6713 & 0.8062 & 0.6596 & 0.0707 & 0.0099 & 0.6319 \\
BERT+CNN & 0.8104 & 0.6666 & 0.8019 & 0.6592 & 0.0724 & 0.0099 & 0.6306 \\
BERT+GRU & 0.8153 & 0.6764 & 0.8097 & 0.6700 & 0.0697 & 0.0092 & 0.6566 \\
\bottomrule
\end{tabular}
}
\end{table}

We conduct a comprehensive comparison between  DyFuLM and several representative baselines, including single-encoder models (BERT, RoBERTa, DeBERTa, DistilBERT, EmoBERTa) and dual-encoder architectures (Simple+Concat, Gated-Fusion, RoBERTa+CNN, BERT+CNN, RoBERTa+BiLSTM, BERT+GRU). As shown in Table \ref{tab2}, DyFuLM consistently achieves superior results across all classification metrics, including Coarse Accuracy, Fine Accuracy, Coarse F1, and Fine F1, indicating a better balance between precision and recall.

In addition to classification performance,  DyFuLM achieves the lowest MAE (0.0674) and MSE (0.0082) values among all models, reflecting higher prediction accuracy in regression tasks. DyFuLM also leads in $R^2$ (0.6903), further emphasizing its strong generalization ability. These results confirm that DyFuLM not only improves overall classification performance but also exhibits strong generalization capabilities in minimizing prediction errors.

\begin{figure}[ht]
  \centering
  \includegraphics[width=1.0\linewidth]{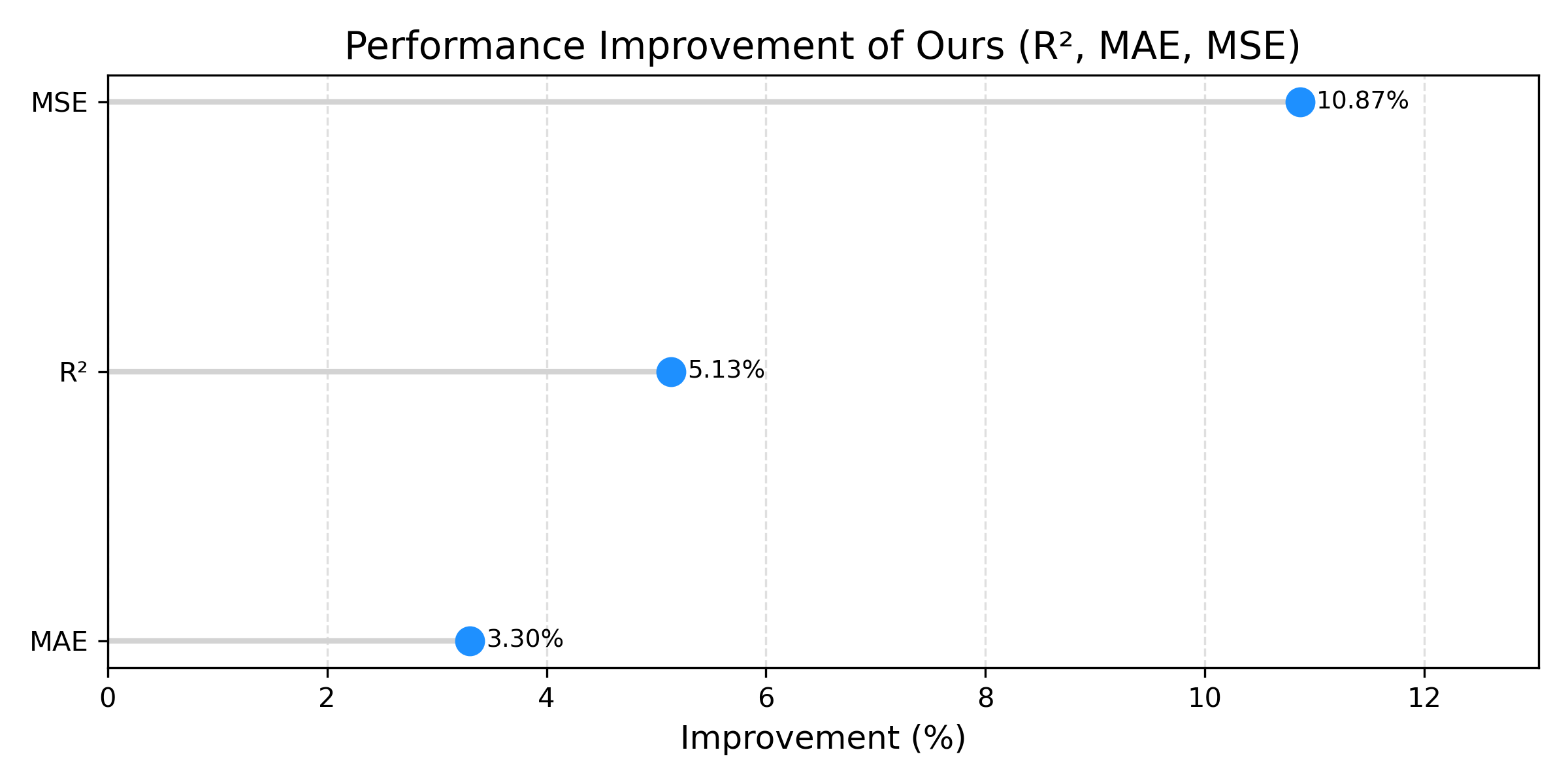}
 \caption{Percentage of Performance Improvement}
 \label{fig5}
\end{figure}
Fig.\ref{fig5} compares DyFuLM with the strongest baseline on the three regression metrics.
First, MSE drops by 10.87\%, indicating a substantial reduction in overall error. Second, the $R^2$ score rises by 5.13\%, showing that the model explains a larger share of the variance in the target values. Finally, MAE decreases by 3.30\%, confirming tighter point-wise predictions. Together, the simultaneous increase in $R^2$ and the decreases in both MAE and MSE demonstrate that the proposed model delivers more accurate and more reliable estimates across the full data range.

\subsection{Ablation Study}
To assess the effectiveness of each component in our hybrid multi-task sentiment analysis framework, we perform ablation studies by individually removing each module while keeping all other settings constant (learning rate = 1e-5, batch size = 16, epochs = 6). All experiments are performed under identical hardware conditions, using an Intel Core i9-13900K CPU, an NVIDIA RTX 4090D GPU, and 64 GB RAM. This controlled setup ensures that performance variations are solely attributed to the excluded module, providing a clear and fair evaluation of its contribution.

\begin{table}[ht]
\centering
\caption{Ablation study of model components (Values are presented to four decimal places, with optimal results for each metric in bold)}
\label{tab3}
\begin{large}
\resizebox{1.0\columnwidth}{!}{
\begin{tabular}{p{4.8cm}ccccccc}
\toprule
\textbf{Experiments} & \textbf{Coarse Acc} & \textbf{Coarse F1} & \textbf{Fine Acc} & \textbf{Fine F1} & \textbf{MAE} & \textbf{MSE} & \textbf{R²} \\
\midrule
\textbf{DyFuLM} & \textbf{0.8264} & \textbf{0.8215} & \textbf{0.6848} & \textbf{0.6814} & \textbf{0.0674} & \textbf{0.0082} & \textbf{0.6903} \\
\makecell[l]{w/o GF+HG+DL} & 0.8173 & 0.8129 & 0.6780 & 0.6729 & 0.0715 & 0.0100 & 0.6277 \\
\makecell[l]{w/o HG+DL} & 0.8189 & 0.8147 & 0.6793 & 0.6769 & 0.0707 & 0.0095 & 0.6439 \\
\makecell[l]{w/o DL} & 0.8186 & 0.8149 & 0.6822 & 0.6790 & 0.0686 & 0.0094 & 0.6507 \\

\bottomrule
\end{tabular}
}
\end{large}
\end{table}

\begin{figure}[ht]
  \centering
  \includegraphics[width=1.0\linewidth]{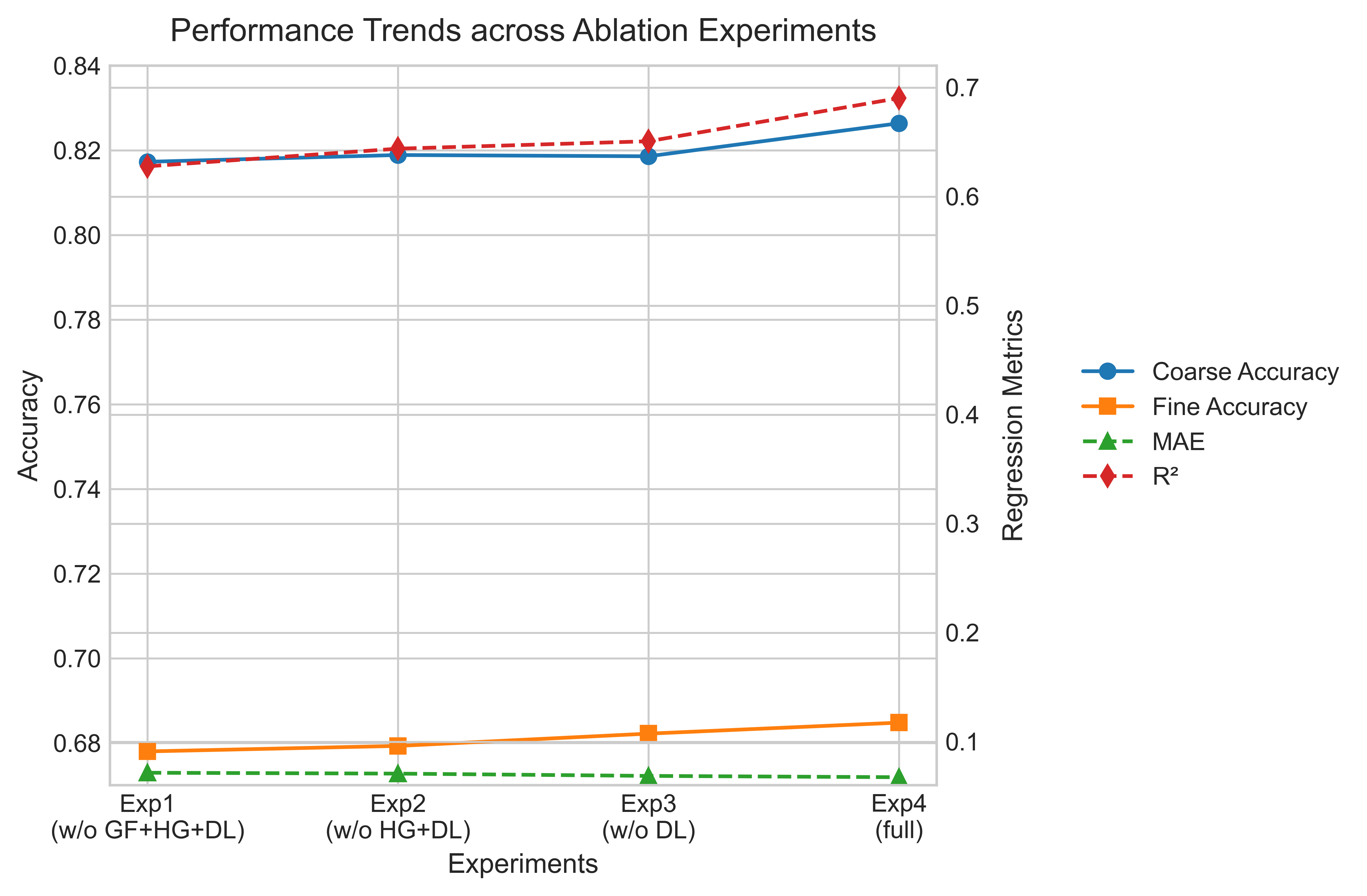}
 \caption{Performance Trends across Ablation Experiments }
 \label{fig6}
\end{figure}

As shown in Table \ref{tab3}, we conduct an ablation study to quantify the effect of each module. The first variant removes all modules, eliminating gated fusion, hierarchical guidance, and dynamic loss weighting. The second variant keeps only the gated fusion module to examine its independent contribution. The third variant removes the dynamic loss weighting while preserving other components. The full model, DyFuLM, integrates all modules. This stepwise comparison clearly demonstrates how each design choice contributes to the overall performance.

The ablation results underscore the effectiveness of each proposed component. The complete DyFuLM model achieves the best overall results, with coarse-grained and fine-grained accuracies of 82.64\% and 68.48\%, respectively, along with the lowest MAE and MSE and the highest R². When all modules are removed, the performance drops by 0.91\% and 0.68\%, highlighting the importance of module cooperation. Furthermore, retaining only the gated fusion module or removing the dynamic loss weighting mechanism leads to additional declines of 0.75\% / 0.55\% and 0.78\% / 0.26\%, respectively, indicating that both modules play key roles in feature interaction and task balance. Overall, the ablation results confirm the effectiveness of each component and demonstrate that their synergy significantly enhances DyFuLM’s semantic representation and sentiment analysis stability.

The ablation results in Fig.\ref{fig6} confirm the effectiveness of each proposed component. Among them, the gated fusion module yields the most substantial performance gain, followed by the hierarchical guidance and dynamic layer extraction modules. Together, these components form a unified framework that enhances both hierarchical understanding and fine-grained sentiment recognition, while also improving training stability.

\section{Conclusion}
In this study, we introduced DyFuLM, a multimodal framework for multi-dimensional sentiment analysis. By capturing both hierarchical semantics and fine-grained emotional cues, DyFuLM achieves this. The proposed hierarchical dynamic fusion and gated feature aggregation modules enable adaptive alignment across layers and models, improving contextual representation and sentiment understanding.

Comprehensive experiments demonstrated that DyFuLM consistently outperforms baseline models in tasks. It achieves higher coarse and fine-grained accuracy while maintaining lower MAE and MSE, indicating a strong balance between expressiveness and precision. Ablation analyses confirmed that the gated fusion and hierarchical guidance modules are key to enhancing robustness and accuracy.

Despite these strengths, DyFuLM has several limitations. It currently focuses on textual input and has not incorporated multimodal signals such as images or audio. Moreover, its multimodal structure increases computational cost, which may limit scalability in real-time applications. In addition, evaluation has so far been limited to a specific domain, and broader validation is required to test its general applicability.

In future work, we plan to extend DyFuLM in three main directions. We will first integrate visual and acoustic modalities to develop multimodal sentiment analysis, enabling the model to capture richer emotional cues beyond text. Next, we aim to enhance generalization and robustness through cross-lingual and cross-domain transfer learning, addressing variations across datasets and contexts. Finally, we will improve computational efficiency by exploring lightweight architectures and model compression for practical deployment. These future efforts will further strengthen DyFuLM’s adaptability and comprehensive understanding of human emotions in real-world scenarios.

\section*{Acknowledgment}
This work was conducted at Xi’an Jiaotong-Liverpool University. The authors would like to thank the university for providing laboratory facilities and technical support. The study also benefited from the use of the Booking.com dataset.

\vspace{12pt}

\end{document}